\pdfoutput=1
\documentclass[11pt,a4paper]{article}
\usepackage[hyperref]{acl2018}
\usepackage{times}
\usepackage{latexsym}

\usepackage{amsmath}
\usepackage{url}
\usepackage{multirow}
\usepackage{tikz-dependency}

\aclfinalcopy 
\def\aclpaperid{419} 

\title{Parser Training with Heterogeneous Treebanks}

\author{Sara Stymne, Miryam de Lhoneux, Aaron Smith, and Joakim Nivre \\
  Department of Linguistics and Philology\\
  Uppsala University\\
  {\tt firstName.lastName@lingfil.uu.se}}
  
\date{}

\begin{document}
\maketitle
\begin{abstract}
How to make the most of multiple heterogeneous treebanks when training a monolingual dependency parser is an open question. We start by investigating previously suggested, but little evaluated, strategies for exploiting multiple treebanks based on concatenating training sets, with or without fine-tuning. We go on to propose a new method based on treebank embeddings. We perform experiments for several languages and show that in many cases fine-tuning and treebank embeddings lead to substantial improvements over single treebanks or concatenation, with average gains of 2.0--3.5 LAS points. We argue that treebank embeddings should be preferred due to their conceptual simplicity, flexibility and extensibility. 
\end{abstract}

\section{Introduction}

In this paper we 
investigate how to train monolingual parsers in the situation where several treebanks are available for a single language. This is quite a common occurrence; in release 2.1 of the Universal Dependencies (UD) treebanks \cite{ud2_1}, 25 languages have more than one treebank. These treebanks can differ in several respects: they can contain material from different language variants, domains, or genres, and written or spoken material. Even though the UD project provides guidelines for consistent annotation, treebanks can still differ with respect to annotation choices, consistency and quality of annotation. 
Some treebanks are thoroughly checked by human annotators, whereas others are based entirely on automatic conversions. All this means that it is often far from trivial to combine multiple treebanks for the same language.

The 2017 CoNLL Shared Task on Universal Dependency Parsing \cite{zeman-EtAl:2017:K17-3} included 15 languages with multiple treebanks. An additional parallel test set of 1000 sentences, PUD, was also made available for a selection of languages. 
Most of the participating teams did not take advantage of the multiple treebanks, however, and simply trained one model per treebank instead of one model per language. There were a few exceptions to this rule, but these teams typically did not investigate the effect of their proposed strategies in detail.

In this paper we begin by performing a thorough investigation of previously proposed strategies for training with multiple treebanks for the same language. We then propose a novel method, based on treebank embeddings. Our new technique has the advantage of producing a single flexible model for each language, regardless of the number of treebanks. We show that this method leads to substantial improvements for many languages. Of the competing methods, training on the concatenation of treebanks, followed by fine-tuning for each treebank, also performed well, but this method results in longer training times and necessitates multiple unwieldy models per language.

\section{Training with Multiple Treebanks}

The most obvious way to combine treebanks for a particular language, provided that they use the same annotation scheme, 
is simply to concatenate the training sets. 
This has the advantage that it does not require any modifications to the parser itself, and it produces a single model that can be directly used for any input from the language in question.  
\newcite{bjorkelund-EtAl:2017:K17-3} and \newcite{das-zaffar-sarkar:2017:K17-3} used this strategy to parse the PUD test sets in the 2017 CoNLL Shared Task. Little details are given on the results, but while it was successful on dev data for most languages, results were mixed on the actual PUD test sets. 
For the two Norwegian language variants, concatenation has been proposed \cite{velldal+2017NoDaLiDa}, but it hurts results unless combined with machine translation.

Training on concatenated treebanks can be improved by a subsequent fine-tuning step.
In this set-up, after training the model on concatenated data, it is refined for each treebank by training only on its own training set for a few additional epochs. This enables the models to learn differences between treebanks, but it requires more training, and results in separate models for each treebank. When the parser is applied to new data, there is thus a choice of which fine-tuned version to use. This approach was used by \newcite{che-EtAl:2017:K17-3} and \newcite{shi-EtAl:2017:K17-3} for languages with multiple treebanks in the CoNLL 2017 Shared Task. 
\newcite{che-EtAl:2017:K17-3} apply fine-tuning to all but the largest treebank for each language, and show average gains of 1.8 LAS for a subset of nine treebanks.
\newcite{shi-EtAl:2017:K17-3} show that the choice of treebank for parsing the PUD test set is important, but do not have any specific evaluation of the effect of fine-tuning.

Another approach, not explored in this paper, is shared gated adversarial networks,
proposed by \newcite{sato-EtAl:2017:K17-3} for the CoNLL 2017 Shared Task. They use treebank prediction  as an adversarial task. 
In this model, treebank-specific BiLSTMs are constructed for all treebanks in addition to a shared BiLSTM which is used both for parsing and for the adversarial task. This method requires knowing at test time which treebank the input belongs to.  \newcite{sato-EtAl:2017:K17-3} show that this strategy can give substantial improvements, especially for small treebanks. For large treebanks, however, there are mostly no or only minor improvements. 

Our approach for taking advantage of multiple treebanks is to use a treebank embedding to represent the treebank to which a sentence belongs. In our proposed model, all parameters of the model are shared; the treebank embedding facilitates soft sharing between treebanks at the word level, and allows the parser to learn treebank-specific phenomena. 
At test time, a treebank identifier has to be given for the input data. 
A key benefit of using treebank embeddings is that we can train a single model for each language using all available data while remaining sensitive to the differences between treebanks. 
The addition of treebank embeddings requires only minor modifications to the parser (see section \ref{the_parser}). 
To the best of our knowledge this approach is novel when applied to the monolingual case as treebank embeddings. The most similar approach we have found in the literature is \newcite{lim-poibeau:2017:K17-3}, who used one-hot treebank representations to combine data 
for improving monolingual parsing for three tiny treebanks, with improvements of 0.6--1.9 LAS.
It is also related to work on domain embeddings for machine translation \cite{kobus+17mtDomain}, and language embeddings for parsing \cite{TACL892}.

We previously used a similar architecture for
combining languages with very small training sets with additional languages \cite{delhoneux-EtAl:2017:K17-3}. 
Language embeddings have also been explored for other cross-lingual tasks such as language modeling \cite{tsvetkov-EtAl:2016:N16-1,ostling-tiedemann:2017:EACLshort} and POS-tagging \cite{bjerva-augenstein:2018:IWCLUL}.
Cross-lingual parsing, however, often requires substantially more complex models. They typically include features such as  multilingual word embeddings \cite{TACL892}, linguistic re-write rules \cite{aufrant+16}, or machine translation \cite{tiedemann:2015:Depling}. Unlike much work on cross-lingual parsing, we do not focus on a low-resource scenario.

\section{Experimental Setup}

We perform experiments for 24 treebanks from 9 languages, using 
UUParser \citep{delhoneux-EtAl:2017:K17-3,delhoneux17arc}. 
We compare concatenation (\textsc{concat}), concatenation with fine-tuning (\textsc{c+ft}), and treebank embeddings (\textsc{tb-emb}). In addition we compare these results to using only single treebanks for training (\textsc{single}). While some of these methods were previously suggested in the literature, no proper evaluation and comparison between them has been performed. For the PUD test data, there is no corresponding training set, so we need to choose a model or set a treebank embedding based on some other treebank. We call this a \textit{proxy} treebank. 

For evaluation we use labeled attachment score (LAS). Significance testing is performed using a randomization test, with the script from the CoNLL 2017 Shared Task.\footnote{\url{https://github.com/udapi/udapi-python/blob/master/udapi/block/eval/conll17.py}}

\subsection{The Parser}
\label{the_parser}

We use UUParser\footnote{\url{https://github.com/UppsalaNLP/uuparser}} \cite{delhoneux-EtAl:2017:K17-3}, which is based on the transition-based parser of \citet{kiperwasser16}, and adapted to UD. It uses the arc-hybrid transition system from \citet{kuhlmann11} extended with a \textsc{Swap} transition and a static-dynamic oracle, as described in \citet{delhoneux17arc}. This model allows the construction of non-projective dependency trees \cite{nivre09acl}.  

A configuration $c$ is represented by a feature function $\phi(\cdot)$ over a subset of its elements and, for each configuration, transitions are scored by a classifier. In this case, the classifier is a multi-layer perceptron (MLP) and $\phi(\cdot)$ is a concatenation of the \mbox{BiLSTM} vectors $v_i$ of words on top of the stack and at the beginning of the buffer. The MLP scores transitions together with the arc labels for transitions that involve adding an arc.

For an input sentence of length $n$ with words $w_1,\dots,w_n$, the parser creates a sequence of vectors $x_{1:n}$, where the vector $x_i$ representing $w_i$ is the concatenation of a word embedding $e(w_i)$ and a character vector, obtained by running a \mbox{BiLSTM} over the $m$ characters $ch_1,\dots,ch_m$ of $w_i$:
\begin{align*}
    x_i &= e(w_i) \circ \textsc{BiLstm}(ch_{1:m})
\end{align*}    
Note that no POS-tags or morphological features are used in this parser.

In the \textsc{tb-emb} setup, we also concatenate a treebank embedding $tb(w_i)$ to the representation of $w_i$:
\vspace{-1mm}
\begin{align*}
    x_i &= e(w_i) \circ \textsc{BiLstm}(ch_{1:m}) \circ tb(w_i)
\end{align*}
Finally, each input element is represented by a \mbox{BiLSTM} vector, $v_i$:
\vspace{-1mm}
\begin{align*}
    v_i &= \textsc{BiLstm}(x_{1:n},i)
\end{align*}

All embeddings are initialized randomly, and trained together with the \mbox{BiLSTMs} and MLP. 
For hyperparameter settings we used default values from \newcite{delhoneux-EtAl:2017:K17-3}. The dimension of the treebank embedding is set to 12 in our experiments; we saw only small and inconsistent changes when varying the number of dimensions.
We train the parser for 30 epochs per setting. For \textsc{c+ft} we apply fine-tuning for an additional 10 epochs for each treebank. We pick the best epoch based on LAS score on the dev set, using average dev scores when training on more than one treebank, and apply the model from this epoch to the test data.

\subsection{Data}

We performed all experiments on UD version 2.1 treebanks \cite{ud2_1}, using gold sentence and word segmentation. We selected 9 languages, based on the criteria that they should have at least two treebanks with fully available training data and a PUD test set.
The sizes of the training corpora for the 9 languages are shown in Table \ref{tab:res}.
The situation is quite different across languages with either treebanks of roughly the same size, as for Spanish, or very skewed data sizes with a mix of large and small treebanks, as for Czech. 
In all cases we use all available data, except for Czech, where we randomly choose a maximum of 15,000 sentences per treebank per epoch for efficiency reasons.

\begin{table*}[tb]
  \centering
  \begin{small}
    \begin{tabular}{l l r | l l l l | l l l l} \hline
 & &  & \multicolumn{4}{c|}{Same treebank test set} & \multicolumn{4}{c}{PUD test set}  \\ 
    \textbf{Language} & \textbf{Treebank} & \textbf{Size} & \textsc{single} & \textsc{concat} & \textsc{c+ft} & \textsc{tb-emb} & \textsc{single} & \textsc{concat} & \textsc{c+ft} & \textsc{tb-emb}  \\ \hline
    \multirow{4}{*}{Czech} & PDT & 68495 & 86.7  & 87.5$^+$  & \textbf{88.3}$^*$  & 87.2$^+$  & \textbf{81.7}  & \multirow{4}{*}{\textbf{81.7}}  & 81.6 & 81.2 \\
           & CAC & 23478 &86.0  & 87.8$^+$  & 88.1$^+$  & \textbf{88.5}$^+$  & 75.0  &  & 81.3 & 81.1 \\
                      &FicTree  & 10160 & 84.3  & 89.3$^+$  & \textbf{89.5}$^+$  & 89.2$^+$  & 66.1  &  & 79.8 & 80.3 \\
     & CLTT & 860& 72.5  & 86.2$^+$  & \textbf{86.9}$^+$  & 86.0$^+$   & 42.1  &  & 80.8 & 80.9 \\ \hline
    \multirow{3}{*}{English} & EWT & 12543 & 82.2  & 82.1  & 82.5   & \textbf{83.0}  & 80.7  & \multirow{3}{*}{80.0} & 81.7$^*$ & \textbf{81.9}$^*$ \\
           &  LinES& 2738 &72.1  & 76.7$^+$  & \textbf{77.3}$^+$  & \textbf{77.3}$^+$  & 62.6  &  & 75.9 & 74.5 \\
           & ParTUT & 1781 & 80.5  & 83.5$^+$  & 85.4$^+$  & \textbf{85.7}$^+$  & 68.0  &  & 78.1 & 76.9 \\ \hline
    \multirow{2}{*}{Finnish} & FTB & 14981& 76.4$^\times$   & 74.4  & 80.1$^*$  & \textbf{80.6}$^*$  & 46.7  & \multirow{2}{*}{73.0} & 54.6 & 53.1 \\
             & TDT & 12217 & 78.1$^\times$  & 70.6  & \textbf{80.6}$^*$  & 80.3$^*$  & 78.6$^\times$  &  & \textbf{81.3}$^*$ & 80.9$^*$ \\ \hline
    \multirow{4}{*}{French}            & FTB & 14759 &83.2 & 83.2  & 83.9$^*$  & \textbf{84.1}$^*$  & 72.0  & \multirow{4}{*}{79.4} & 76.7 & 74.1 \\
                      & GSD & 14554 & 84.5  & 84.1  & 85.3  & \textbf{85.6}$^\times$  & 79.1  &  & 80.2$^*$ & \textbf{80.3}$^*$ \\  
                      & Sequoia& 2231 & 84.0  & 86.0$^+$  & \textbf{89.8}$^*$  & 89.1$^*$   & 69.5  &  & 78.1 & 77.6 \\ 
                      & ParTUT & 803 & 79.8  & 80.5  & 89.1$^*$  & \textbf{90.3}$^*$  & 63.4  &  & 78.8 & 77.5 \\   \hline 
    \multirow{3}{*}{Italian} & ISDT & 12838 & 87.7  &  \textbf{87.9} & 87.7  & 87.6  & 85.4  & \multirow{3}{*}{86.0} & 85.7 & 86.0 \\
            & PoSTWITA & 2808 & 71.4  & 76.7$^+$  & 76.8$^+$  & \textbf{77.0}$^+$  & 68.5  &  & 85.7 & 85.3 \\
            & ParTUT & 1781 &  83.4 & 89.2$^+$  & \textbf{89.3}$^+$  & 88.8$^+$  & 77.4  &  & 85.8$^+$ & \textbf{86.1}$^+$ \\ \hline
    \multirow{2}{*}{Portuguese} 
                & GSD & 9664 & 88.3  & 87.3  & 89.0$^*$  & \textbf{89.1}$^*$  & 74.0  & \multirow{2}{*}{76.8$^+$} & 75.2 &74.9  \\
                & Bosque & 8331 & 84.7  & 84.2  & 86.2$^\times$  & \textbf{86.3}$^*$  & 75.2  &  & 77.5$^+$ & \textbf{77.6}$^+$  \\ \hline
    \multirow{2}{*}{Russian} & SynTagRus & 48814 & 90.2$^\times$  & 89.4  & \textbf{90.4}$^\times$  & \textbf{90.4}$^\times$  & 66.0  & \multirow{2}{*}{68.7} & 66.3 & 66.4 \\
                  & GSD & 3850 & 74.7$^\times$  & 73.4  & 79.8$^*$  & \textbf{80.8}$^*$  & 70.1$^\times$  &  & 77.6$^*$ & \textbf{78.0}$^*$ \\   \hline
    \multirow{2}{*}{Spanish} & AnCora & 14305 & 87.2$^\times$  & 86.2  & 87.5$^\times$  & \textbf{87.6}$^\times$  & 75.2  &\multirow{2}{*}{79.9}  & 77.7 & 76.4 \\
               & GSD & 14187 & 84.7  & 83.0  & 85.8$^\times$  & \textbf{86.2}$^*$  & 79.8  &  & 80.8$^+$ &  \textbf{80.9}$^*$ \\ \hline
    \multirow{2}{*}{Swedish} &  Talbanken & 4303 & 79.6  & 79.1  &  80.2 & \textbf{80.6}$^\times$  & 70.3  & \multirow{2}{*}{72.0$^+$} & 73.2$^*$ & \textbf{73.6}$^*$ \\
                      &  LinES& 2738 &74.3  & 76.8  & \textbf{77.3}$^+$  & 77.1$^+$  & 64.0  &  & 70.0 &  69.0 \\ \hline
 Average & &  &81.4 & 82.7$^+$ & \textbf{84.9}$^*$ & \textbf{84.9}$^*$  & 77.9  & 77.5 & 80.0$^*$ & \textbf{80.1}$^*$ \\  
\hline
  \end{tabular}
\end{small}
\caption{LAS scores when testing on the training treebank and on the PUD test set with training treebank as proxy. For each test set, the best result is marked with bold. Treebank size is given as number of sentences in the training data. Statistically significant differences, at the 0.05-level,  from \textsc{single} are marked with +, from \textsc{concat} with $\times$ and from both these systems with *. For clarity,  significance for PUD is only shown for the proxy treebank with the highest score.}
  \label{tab:res}
\end{table*}

\begin{figure*}[t]
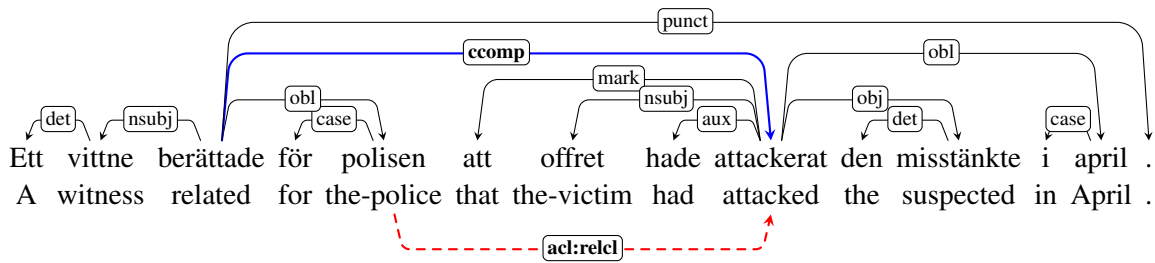

\centering
\begin{dependency}
\begin{deptext}
Ett \& vittne \& ber{\"a}ttade \& f{\"o}r \& polisen \& att \& offret \& hade \& attackerat \& den \& misst{\"a}nkte \& i \& april \& . \\[1mm]
A \& witness \& related \& for \& the-police \& that \& the-victim \& had \& attacked \& the \& suspected \& in \& April \& . \\  
\end{deptext}
\depedge[edge unit distance=0.7em]{2}{1}{det}
\depedge[edge unit distance=0.7em]{3}{2}{nsubj}
\depedge[edge unit distance=0.7em]{5}{4}{case}
\depedge[edge unit distance=0.7em]{3}{5}{obl}
\depedge[edge unit distance=0.7em]{9}{6}{mark}
\depedge[edge unit distance=0.7em]{9}{7}{nsubj}
\depedge[edge unit distance=0.7em]{9}{8}{aux}
\depedge[edge unit distance=0.5em, edge style={thick,blue}]{3}{9}{\textbf{ccomp}}
\depedge[edge below, edge style={dashed,thick,red}, edge unit distance=0.3em]{5}{9}{\textbf{acl:relcl}}
\depedge[edge unit distance=0.7em]{11}{10}{det}
\depedge[edge unit distance=0.7em]{9}{11}{obj}
\depedge[edge unit distance=0.7em]{13}{12}{case}
\depedge[edge unit distance=0.75em]{9}{13}{obl}
\depedge[edge unit distance=0.37em]{3}{14}{punct}
\end{dependency}
\caption{Example sentence from the Swedish PUD treebank with parsing error represented by dashed arc. Translation: ``A witness told the police that the victim had attacked the suspect in April.''}
\label{fig:error}
\end{figure*}
\section{Results}

Table \ref{tab:res} shows the results on the test sets of each training treebank and on the PUD test sets. Overall we observe substantial gains when using either \textsc{c+ft} or \textsc{tb-emb}. On average both \textsc{c+ft} and \textsc{tb-emb} beat \textsc{single} by 3.5 LAS points and \textsc{concat} by over 2.0 LAS points when testing on the test sets of the treebanks used for training, and both methods beat both baselines by over 2.0 LAS points for the PUD test set, if we consider the best proxy treebank.  

We see positive gains across many scenarios when using \textsc{c+ft} and \textsc{tb-emb}. First, there are gains for both balanced and unbalanced data sizes, as in the cases of Spanish and French, respectively. Secondly, there are cases with different language variants, as for Portuguese, and different domains, as for Finnish where FTB only contains grammar examples and TDT contains a mix of domains. There are also cases of known differences in annotation choices, as for the Swedish treebanks.

When the data is very skewed, as for Russian, the effect of adding a small treebank to a large one is minor, as expected. 
While our results are not directly comparable to the adversarial learning in \newcite{sato-EtAl:2017:K17-3} who used a different parser and test set, the improvements of \textsc{c+ft} and \textsc{tb-emb} are typically at least on par with and often larger than their improvements. While our improvements are, unsurprisingly, largest for smaller treebanks, we do also see some improvements for large treebanks, in contrast to \newcite{sato-EtAl:2017:K17-3}.

Some variation can be observed between languages. In two cases, Italian ISDT and Czech PUD, \textsc{concat} performs marginally better than the more advanced methods, but these differences are not statistically significant. In several cases, especially for small treebanks, \textsc{concat} helps noticeably over \textsc{single}, whereas it actually hurts for  Finnish and Russian. It is, however, nearly always better to combine treebanks in some way than to use only a single treebank.
The differences between the two best methods, \textsc{c+ft} and \textsc{tb-emb} are typically small and not statistically significant, with the exception of Czech PDT, and for some of the small proxy treebanks for PUD.

The PUD test set can be seen as an example of applying the proposed models to unseen data, without matching training data.
For all languages, except Czech, the results for \textsc{c+ft} and \textsc{tb-emb} with the best proxy treebank are significantly better than the equivalent result for \textsc{single}, and for six of the nine languages,   \textsc{tb-emb} performs significantly better than \textsc{concat}.
It is clear that some treebanks are bad fits to PUD, most notably Finnish FTB and Russian SynTagRus. However, even when a treebank is a bad fit, \textsc{tb-emb} and \textsc{c+ft} can still improve substantially over using only the single model for the treebank with the best fit, as for Russian where there is a gain of nearly 8 LAS points for \textsc{tb-emb} over \textsc{single}, when using GSD as a proxy.
For some languages, however, most notably Italian, the choice of proxy treebank makes little difference for \textsc{tb-emb} and \textsc{c+ft}. It is also interesting to see that in many cases it is not the largest treebank that is the best proxy for PUD. 
The large difference in results for PUD, depending on which treebank was used as proxy, also seems to point at potential inconsistencies in the UD annotation for several languages.

\section{Error Analysis}

To complement the LAS scores, we performed a small manual error analysis for Swedish, looking at the results for the PUD data, when translated using different methods and proxy treebanks. The two Swedish treebanks, Talbanken and LinES, are known to differ in the annotation of a few constructions, notably relative clauses and prepositions that take subordinate clauses as complements. The error analysis reveals that the treebank embedding approach allows the model to learn the distinctive ``style'' of each treebank, while concatenation, even with fine-tuning, results in more inconsistencies in the output. A typical example is shown in Figure 1. When trained with treebank embeddings (and Talbanken as the proxy treebank), the parser produces the correct tree depicted above the sentence. When trained with fine-tuning instead, the parser incorrectly analyzes the subordinate clause as a relative clause (as shown by the dashed arc below the sentence), because the \emph{mark} relation is also used for relative pronouns in the LinES treebank, despite the fact that such structures never occur in Talbanken.

\section{Conclusion and Future Work}

We have conducted the first large-scale study on how best to combine multiple treebanks for a single language, when all treebanks use the same annotation scheme but may be heterogeneous with respect to domain, genre, size, language variant, annotation style, and quality, as is the case for many
languages in the UD project. 
We propose using treebank embeddings, which represent the treebank a sentence comes from. This method is simple, effective, and flexible, and performs on par with a previously suggested method of using concatenation in combination with fine-tuning, which, however,  requires longer training, and produces more models.

We show that both these methods give substantial gains for a variety of languages, including different scenarios with respect to the mix of available treebanks. Our results are also at least on par with a previously proposed, but more complex  model, based on adversarial learning 
\cite{sato-EtAl:2017:K17-3}. To improve parsing accuracy, it is certainly worth combining multiple treebanks, when available, for a language, using more sophisticated methods than simple concatenation. We recommend the treebank embedding model due to its simplicity.

The proposed methods work well with a transition-based parser with BiLSTM feature extractors without POS-tags or pre-trained embeddings. In future work, we want to investigate how these methods interact with other parsers, and if the combination methods are useful also for tasks like POS-tagging and morphology prediction. 

We did not yet investigate methods for choosing a proxy treebank when parsing new data. The results on the PUD test set could indicate which treebank is likely to be the best proxy for the languages explored here. 
Other factors that could be taken into account when making this choice include degree of domain match and treebank quality. 
The user may also simply choose the desired annotation style by selecting the corresponding proxy treebank.
For the \textsc{tb-emb} approach, interpolation of the various treebank embeddings is another possibility.

In the current paper, we explore only the monolingual case, using several treebanks for a single language. Preliminary experiments show that we can combine treebank and language embeddings and add other languages to the mix. Including closely related languages typically gives additional gains, which we will explore in future work.

\section*{Acknowledgments}

We gratefully acknowledge funding from the Swedish Research Council (P2016-01817) and computational resources on the Taito-CSC cluster in Helsinki from NeIC-NLPL (\url{www.nlpl.eu}).

\bibliography{confs-long,refs-basic}
\bibliographystyle{acl_natbib}

\end{document}